\documentclass[11pt]{article}
\usepackage[a4paper,margin=1in]{geometry}
\usepackage{amsmath,amssymb,amsfonts}
\usepackage{bm}
\usepackage{hyperref}
\usepackage{enumitem}
\usepackage{algorithm}
\usepackage{algpseudocode}
\usepackage{graphicx}
\usepackage{mathtools}
\usepackage[sort]{natbib}

\title{\textbf{The Ups and Downs of Backprop Weights}}
\author{
\begin{tabular}{c}
~\\[-0.2em]
\textbf{Giuseppe Chindemi}\\
\textnormal{Institute of Neuroinformatics,}\\
\textnormal{UZH / ETH Zurich}\\
\textnormal{ETH AI Center}\\
\textnormal{Zurich, Switzerland}\\
\texttt{giuseppe.chindemi@ai.ethz.ch}
\end{tabular}
\and
\begin{tabular}{c}
~\\[-0.2em]
\textbf{Benjamin F.~Grewe}$^{*}$\\
\textnormal{Institute of Neuroinformatics,}\\
\textnormal{UZH / ETH Zurich}\\
\textnormal{ETH AI Center}\\
\textnormal{Zurich, Switzerland}\\
\texttt{bgrewe@ethz.ch}
\end{tabular}
}

\newcommand{\vect}[1]{\bm{#1}}
\newcommand{\mat}[1]{\bm{#1}}

\begin{document}
\maketitle
\begin{abstract}
Backpropagation (BP) has driven the remarkable success of modern deep learning by enabling large hierarchical networks to learn complex functions end-to-end. Yet BP does not by itself determine how parameters should be organized to allow the selective reuse and adaptation of the network's functional components. For example, object recognition and motion prediction can depend on overlapping parameter sets, making them difficult to isolate or modify independently. We call this condition \emph{weight entanglement}.

Most modern architectures already contain mechanisms that dynamically select which parts of a network participate in a computation. Nonlinearities and competitive operations gate units, attention selects interactions, and Mixture-of-Experts architectures route inputs to modules. Yet this selection does not ensure that a consistent functional component is implemented by the same identifiable parameter set across samples.

We propose to address this problem via \emph{weight operators}, which encourage the network to learn a set of distinct, reusable functional components and how to compose them at inference time, rather than a single unstructured input--output relationship. That is, for each sample the model first infers the operator composition that forms the required function, then updates only the parameter sets of the selected operators. Vector Networks (VNs) represent a first example of how these principles could be implemented in practice. VNs combine the parameter selection process with local error-driven updates within each layer and demonstrate that learned operators can be reused and recombined for combinations absent from training while updates remain restricted to those parameter sets. This provides a concrete basis for testing \emph{functional parameter identifiability}: whether operators continue to implement the same components during learning. We argue that functional parameter identifiability may provide a missing organizing principle for building models that can systematically reuse and recombine learned functions while selectively adapting only what needs to change.
\end{abstract}

\section{Introduction}
\label{sec:introduction}
Modern machine learning owes much of its success to backpropagation of error (BP) \citep{werbos1974beyond,linnainmaa1976taylor,rumelhart1986learning}. By transporting output errors backward through many layers, BP made deep networks trainable end-to-end under a single objective. Transformers \citep{vaswani2017attention} later showed how far the same machinery could be pushed by scaling model size, data, and compute. Across these developments, the core recipe remained remarkably uniform: define a global loss, compute gradients from outputs back to inputs, and optimize all of the network's parameters end-to-end.

However, optimizing network behavior does not by itself organize the learned parameters for selective reuse or revision. Depending on the input, the same parameter may contribute to many different computations, such as object recognition, language processing, or motor control. Such sharing can be useful because it lets networks exploit common structure and generalize across examples and tasks. But when different parts of a function depend on overlapping parameter sets, the parameter set implementing each part cannot be reliably identified, selected, or modified independently.

We call this structural condition \emph{weight entanglement}. To state what overcoming it requires, we use \emph{function} for the complete processing required by a sample or task. A \emph{functional component}, or subfunction, is a reusable part of that function that transforms some or all of an input and contributes to an output representation. Multiple functional components can be composed to form the sample-specific or task-specific function. Overcoming weight entanglement does not require eliminating useful parameter sharing; it requires each functional component to remain linked to an identifiable parameter set across contexts, so that the component can be recruited and its parameter set modified selectively. The ultimate design goal is \emph{addressable reuse}.

An electric motor can illustrate this terminology more concretely. Its overall function is composed of functional components: bearings reduce friction, actuators generate motion, sensors measure state, controllers regulate speed, and couplings transfer force. Each component is a distinct subfunction and has an identifiable physical implementation analogous to an identifiable parameter set. A worn component can therefore be located and replaced, while torque, damping, or speed can be improved by modifying the relevant components. Imagine instead that every physical part is entangled and contributed to every subfunction. No component could then be isolated or modified independently. Repairing or extending such a part-entangled machine would require retuning or rebuilding the whole motor, without any guarantee that unrelated parts of its function would remain intact.

Weight entanglement creates the corresponding problem in a neural network: recombining or changing one learned functional component may alter others because the parameter set implementing it cannot be addressed separately. This limitation becomes clearest in three settings. Compositional out-of-distribution (OOD) generalization asks whether familiar components can be combined into the function required by a new sample when that combination was absent from training. Continual adaptation asks whether the parameter set implementing one component can change while unrelated components remain stable. Model editing asks whether a specific output or learned fact can be corrected without broad collateral effects. Although their causes differ, together these settings test whether learned functional components remain reusable and selectively addressable.

A necessary first step toward addressable reuse is to determine which parts of the network participate in processing each sample. ReLU-like activation functions, attention mechanisms, and MoE routing already provide such selection by choosing units, interactions, or modules. However, these mechanisms do not require the same identifiable parameter set to implement a consistent functional component across samples. We use \emph{functional parameter identifiability} to denote the ability to link a functional component to the parameter set that implements it, with this link remaining stable across contexts and causally testable.

The central challenge is to identify the parameter sets implementing the functional components required by each sample, restrict learning to those sets, and preserve their roles across contexts. One promising family of candidates is local error-driven learning, which uses activities and error signals available within a layer to identify and restrict the learning targets. However, a local update rule does not by itself ensure functional parameter identifiability: the parameter set implementing a functional component must remain identifiable and continue to implement that component across contexts. For example, the parameter set implementing motion prediction should retain the same subfunction across different scenes, so correcting a motion error does not disrupt object recognition.

\section{From conditional participation to functional parameter identifiability}
\label{sec:functional_parameter_identifiability}
Participation, functional parameter identifiability, and plasticity answer three different questions. \emph{Participation} asks which units and parameter sets contribute to the computation for a sample. A parameter set can participate without having a specific functional role because it may contribute jointly to several parts of the function. \emph{Functional parameter identifiability} asks whether a particular functional component can be linked to an identifiable parameter set. This link should remain stable across contexts and be causally selective: ablating that parameter set should remove or impair the corresponding component of the overall function while leaving unrelated components largely intact. \emph{Plasticity} asks which parameter sets are allowed to change during learning.

For example, the function required to process a moving object may combine an object-identification component with a motion-estimation component. The parameter sets implementing both components participate in processing the sample. If an identifiable parameter set implements motion estimation, ablating it should selectively impair motion estimation while preserving object identification. A motion error should therefore update that parameter set without changing the parameter set implementing object identification. BP often links participation directly to plasticity because gradient flow makes parameter sets on participating pathways eligible for updates. It does not first establish which subfunction needs correction or identify the parameter set that implements it. We therefore argue that a learning system needs a mechanism that links participation to plasticity by identifying the parameter set implementing the functional component that requires correction.

Functional parameter identifiability can be understood as a form of \emph{parameter disentanglement}, but it addresses a different question from \emph{representational disentanglement}. It asks whether a functional component remains linked to an identifiable parameter set, whereas representational disentanglement asks whether latent variables align with statistically independent or human-interpretable representational factors through sparsity, group structure, or other priors on activations \citep{locatello2019challenging,locatello2019fairness}. An interpretable latent code does not, by itself, identify the parameter sets implementing the transformations that produce or use it, or show that these sets can be modified selectively. Dense state representations, conversely, can in principle be produced by structured, addressable functional components. Disentangling representations therefore does not establish functional parameter identifiability. Nor does evidence that features share representational directions, on its own, establish which parameter sets implement the transformations that produce them \citep{elhage2022toy}. For example, in a standard BP-trained weight matrix, multiple functional components can depend on overlapping parameter directions without an explicit decomposition into identifiable parameter sets. When this overlap prevents selective access, a change meant to improve one behavior can affect others \citep{goodfellow2015empirical}.

A useful design goal is therefore \emph{structured parameter sharing}, in which the same functional component is implemented by the same identifiable parameter set across contexts, while distinct components remain independently addressable for recruitment, testing, or change. Neural networks already have one ingredient for such targeted change: they can select which units, interactions, or modules participate in processing each sample. Rectified nonlinearities silence units whose pre-activations fall below threshold, so only part of a layer contributes on a given sample. Competitive operations such as max pooling propagate selected responses while suppressing alternatives. Attention selects interactions among representations, while Mixture-of-Experts (MoE) architectures route tokens or samples toward a restricted subset of experts \citep{shazeer2017outrageously,fedus2022switch}. Parameter-efficient methods such as adapters and LoRA confine some later updates to designated parameter subsets \citep{houlsby2019parameterefficient,hu2021lora}. Selective participation can also gate learning: inactive units and branches receive no local gradient, and a routed expert is updated only on its assigned inputs. Yet selective and limited BP updates do not establish an identifiable parameterized implementation for each reusable functional component; one expert can implement several entangled subfunctions, and an adapter need not identify the component it changes. The next step is to make each reusable component an addressable parameterized unit. A sample-specific function should be formed by composing these units, and only the parameter sets implementing the selected components should be updated. The model would thereby select both the functional components used to process a sample and the parameter sets allowed to change when learning from it.

Beyond changes to network architectures and training algorithms, standard benchmark metrics alone cannot assess functional parameter identifiability. High accuracy can result from broad training coverage without a stable decomposition into functional components and their implementing parameter sets. The same component should be implemented by largely the same parameter set across contexts, even when unrelated aspects of the input change. A different required component should alter the selected parameter sets predictably. Selection shows which parameterized units are used; their roles must be tested by measuring the effects of ablating or updating them. Ablating the implementation of a component should mainly impair functions that use it. Updating its parameter set should change the targeted behavior with limited effects on functions that do not use that component. Across continual-learning phases, parameter sets unrelated to the current function should remain untouched and thus comparatively stable. Tests should also check whether a routing change causes other inputs to use different parameter sets even when their values remain fixed. Together, these interventions test whether each parameter set continues to implement the same reusable functional component across contexts.

\section{Functional parameter identifiability enables reuse and selective change}
\label{sec:reuse_change}
Compositional OOD generalization and continual adaptation provide two complementary tests of learned structure. Compositional OOD generalization tests \emph{selective reuse}: familiar functional components must be composed into new functions. Continual adaptation tests \emph{selective change}: the parameter set implementing one component must be modified while unrelated components remain stable. Both are difficult when a well-performing model lacks reusable, addressable functional components.

Compositional OOD benchmarks make the reuse problem explicit. Sequence models trained with BPTT can perform well on familiar combinations yet struggle when known primitives appear in unseen combinations, as shown by SCAN and gSCAN \citep{lake2018generalization,ruis2020benchmark}. ARC poses a related challenge in visual program-like reasoning, where a few demonstrations specify a function that must be inferred and applied to a new configuration \citep{chollet2019measure}. Approaches that introduce explicit search or abstraction over reusable subfunctions offer one route to this problem \citep{alford2022neuralguided}. In embodied and agentic settings, the same demand arises when familiar objects, goals, constraints, and sub-skills appear in combinations not covered during training \citep{brohan2023rt1,ahn2022can}. Together, these settings motivate a direct test of selective reuse: does the model recruit the functional components learned for familiar elements when those components are required in new combinations?

Continual learning poses the complementary problem. Classical work on catastrophic interference showed that sequential gradient updates can overwrite parameter directions on which previous tasks depend \citep{mccloskey1989catastrophic}, and modern continual-learning methods still devote substantial machinery to protecting or replaying earlier knowledge \citep{kirkpatrick2017overcoming,rolnick2019experience,haan2026continual}. Multi-stage training pipelines reveal a related tension when fine-tuning or preference optimization improves one class of behavior while regressing another \citep{ouyang2022training,kotha2024understanding,ghosh2024closer}. Some of these regressions may be failures of participation rather than overwriting. For example, in LLMs translating prompts into other languages recovers part of the capabilities that fine-tuning suppressed, suggesting that the capability survives but is no longer recruited \citep{kotha2024understanding}. Weight entanglement can obscure which parameter sets implement particular functional components and thereby contribute to interference, even though reusing the same functional component across tasks remains desirable. Functional parameter identifiability would not eliminate forgetting, but it could help restrict updates to parameter sets involved in the current function while limiting changes to those used by other functions.

Model editing and unlearning sharpen the same requirement. Correcting a specific output or learned fact, or removing the influence of particular data, is difficult when the relevant function depends on parameter sets shared with other functions. Parameter editing methods must be evaluated for collateral changes, while unlearning may require retraining or additional controls to remove residual information \citep{meng2022locating,meng2023massediting,bourtoule2021machine}. Functional parameter identifiability would make it easier to locate the functional components and their implementing parameter sets before editing the function or removing learned information.

Scaling data and models changes how often the consequences of weight entanglement appear and how severe they are. Larger datasets, models, retrieval systems, tools, and simulation broaden coverage and can reduce the need for online weight updates \citep{kaplan2020scaling,hoffmann2022training,lewis2020retrievalaugmented,schick2023toolformer}. Broader coverage, however, does not imply that the parameter sets implementing learned functional components are easier to identify or modify selectively. This matters particularly in long-tail data distributions \citep{chen2024endtoend}, where rare cases can arise from familiar elements in unfamiliar combinations. In automated driving, corner-case taxonomies include known objects in unusual places or in unusual numbers \citep{breitenstein2020systematization}. Open-ended agent environments make this structure explicit, generating tasks by recombining worlds, goals, co-players and rules \citep{openendedlearningteam2021openended,adaptiveagentteam2023humantimescale}, yet even an agent trained on a pool of billions of such tasks can fail in situations that are rare in its training distribution, such as using one object as a tool to move another \citep{adaptiveagentteam2023humantimescale}. Finding or simulating rare cases can expand training coverage \citep{bogdoll2023one}, but it does not tell the model which learned components to compose into the required function or, when necessary, update.

Robustness, interpretability, and generative confabulation may also benefit from a more addressable organization of parameter sets, but these links are less direct. The strongest motivation here remains selective reuse and selective change of learned functional components and their implementing parameter sets.

\section{From sparse coding to weight operators}
\label{sec:weight_operators}
Sparse coding shows how sparsity and competition can produce reusable, addressable functional components while localizing learning. It represents an input using a small set of learned vectors, called \emph{dictionary elements}. An input $\vect{x}$ is modeled as
\begin{equation}
\vect{x} \approx \mat{S}\,\vect{a},
\end{equation}
where $\mat{S}$ is a learned dictionary and the coefficient vector $\vect{a}$ is constrained to be sparse. Each dictionary element becomes a reusable component of the reconstruction. During inference, the elements compete to explain the input: the support of $\vect{a}$ identifies which components are recruited, and the coefficient values determine how strongly each is used. Because only the recruited elements contribute, learning can update them selectively, allowing their roles to emerge and be refined through repeated use \citep{oja1982simplified,rumelhart1985feature,sanger1989optimal,olshausen1996emergence}. Sparsity thus makes the learned components addressable through the code that selects them. Routed networks likewise select units or modules before an update \citep{shazeer2017outrageously,fedus2022switch}. The question for deep models is whether this principle can extend from dictionary elements to parameterized components that perform reusable transformations across contexts without requiring the state representations themselves to be sparse.

However, simply stacking sparse coding layers to gain depth raises a further issue as the sparse code is also the representation passed to the next layer. Hierarchical sparse-coding and sparse predictive-coding models use this design \citep{rao1999predictive,rozell2008sparse,hyvarinen2001twolayer,mairal2014sparse}. The same activity code selects dictionary elements and carries information forward. Strong sparsity can therefore discard information \citep{rigamonti2011are} or make representations unstable, with similar inputs recruiting different dictionary elements \citep{wang2010localityconstrained}; recognition performance can peak at an intermediate level of sparsity \citep{barak2013sparseness,makhzani2014ksparse}. In contrast, deep networks can learn hierarchical functions while retaining dense, smoothly varying state representations. BP and predictive coding (PC) provide two ways to train such hierarchies. In a conventional multilayer perceptron, each layer applies a learned weighted transformation followed by a nonlinearity. A forward pass computes the output with the weights held fixed, and BP then propagates loss gradients backward through the hierarchy to update the weights. PC instead introduces recurrent \emph{activity inference}: with weights fixed, internal activities are optimized to reconcile observations with predictions exchanged between levels, generating local prediction errors \citep{rao1999predictive}. These errors drive differential Hebbian-style updates that pair local activity with the difference between observed input and the model's prediction. PC is local in the sense that an update uses activity and error signals available within a layer, although those signals can reflect interactions throughout the hierarchy. Both methods can therefore learn hierarchical transformations. Neither learning method, however, establishes functional parameter identifiability by itself. They specify how participating parameters change, but do not establish whether the parameter set implementing a particular component remains identifiable across contexts.

\begin{figure}[t]
\centering
\begin{minipage}[t]{0.50\textwidth}
    \vspace{0pt}
    \centering
    \includegraphics[width=\linewidth]{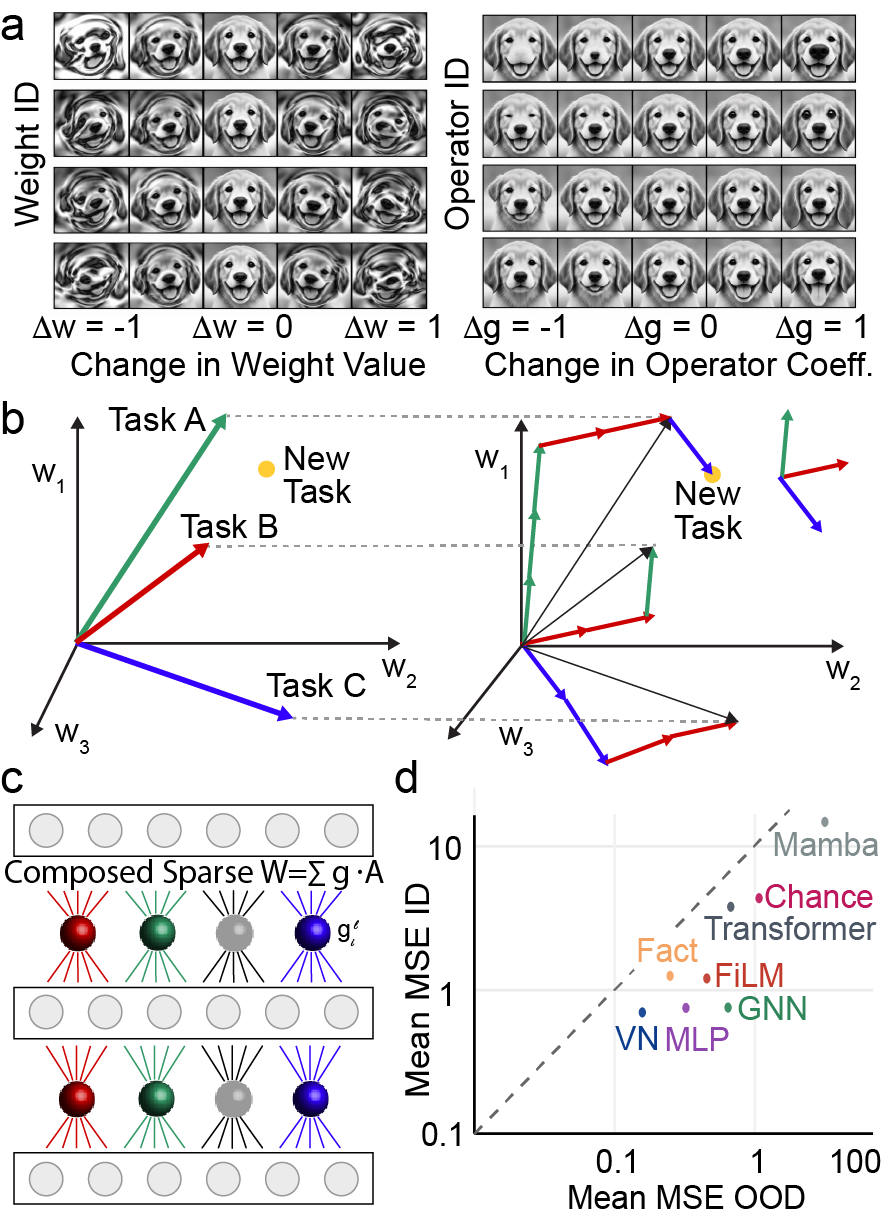}
\end{minipage}
\hfill
\begin{minipage}[t]{0.47\textwidth}
    \vspace{0pt}
    \caption{\textbf{From weight entanglement to functionally identifiable operators.}
    \textbf{(a)} Perturbing individual weights in a conventional network can produce broad, entangled output changes (left; adapted from \citealp{kumar2025questioning}), whereas varying operator coefficients changes distinct features more selectively (right; schematic).
    \textbf{(b)} An entangled parameterization encodes functions in overlapping parameter directions (left); an operator-based representation composes a function from reusable operators, allowing new tasks to recombine them (right; schematic).
    \textbf{(c)} A rank-1 Vector Network (VN) forms an input-specific weight operator from a sparse composition of operator atoms (schematic based on \citealp{pokel2026learning}), $\mat{W}^{\star}(\vect{x})=\sum_k g_k^{\star}(\vect{x})\mat{A}_k$, where $\mat{A}_k=\vect{u}_k\vect{s}_k^\top$ and the inferred coefficients $g_k^{\star}$ determine which atoms participate and are eligible for learning.
    \textbf{(d)} VNs improve compositional out-of-distribution (OOD) generalization over dense and structured baselines on the $n$-body benchmark (adapted from \citealp{pokel2026learning}).}
    \label{fig:weight_entanglement}
\end{minipage}
\end{figure}

Target propagation and bio-inspired deep feedback-control approaches provide another way to train hierarchical networks. Rather than relying only on a backpropagated derivative, they provide local targets or control signals that specify what activity each layer should produce \citep{bengio2014how,lee2015difference,meulemans2021credit}. These signals can guide learning throughout a deep hierarchy, but they do not by themselves identify the reusable functional component requiring correction and the parameter set implementing it.

PC, target propagation, and deep feedback-control approaches thus show that deep hierarchical networks can be trained with layer-specific error, target, and control signals. The remaining requirement is functional parameter identifiability: each reusable functional component must remain linked to an identifiable parameter set that can be selected and updated across contexts while state representations remain dense.

We propose implementing each such component as a \emph{weight operator}: a parameterized functional module that performs a transformation on some or all of an input representation and contributes to an output representation. A rank-1 instance is an \emph{atomic weight operator}.

Figure~\ref{fig:weight_entanglement}a illustrates the intended contrast: perturbing an individual weight in an entangled network can alter several output features, whereas varying the coefficient of an addressable weight operator can produce a more selective transformation. The network recruits and composes such operators into the processing function required for a particular sample (Figure~\ref{fig:weight_entanglement}b). A weight operator need not span a whole layer; its parameter set may define a small matrix, a group of weights, or a subnetwork-like module. The system must be able to infer when the operator is needed, reuse it across contexts, and restrict substantial updates to its parameter set when the component it implements must change. In this formulation, the functional component is the \emph{unit of computation}, its weight operator makes that component identifiable in parameter space, and the operator's parameter set is the \emph{unit of plasticity}. Depending on the functional scale, a component could be implemented by a single weight operator, a stable composition of operators, a pathway, or a temporally extended operator.

Vector Networks (VNs) are energy-based networks that provide one concrete implementation of weight-operator selection and training \citep{pokel2026learning}. A VN whose operator atoms are rank-1 outer products may be called an \emph{Atomic Network}. Within each layer of a VN, fast inference and slower learning optimize the same layer-local energy with respect to different variables: inference changes the operator coefficients, whereas learning changes the selected operators' parameter sets. In the rank-1 formulation considered here, a VN replaces a monolithic inter-layer weight matrix with a dictionary of rank-1 \emph{operator atoms}, each formed by the outer product of an output vector and a sensing vector, which together constitute its parameter set. Each operator atom is a candidate parameterized implementation of a functional component. Let $\vect{x}$ denote the current layer input, $K$ the number of available atoms, and $k$ their index. A subset is recruited for each input (Figure~\ref{fig:weight_entanglement}c). The inferred coefficient vector $\vect{g}^{\star}(\vect{x})=(g_1^{\star}(\vect{x}),\ldots,g_K^{\star}(\vect{x}))^\top$ defines the descriptive input-specific operator
\begin{equation}
\mat{W}^{\star}(\vect{x}) = \sum_{k=1}^{K} g_k^{\star}(\vect{x})\,\mat{A}_k,
\label{eq:vn_effective_operator}
\end{equation}
where $\{\mat{A}_k\}_{k=1}^{K}$ are the learned operator atoms and the superscript $\star$ denotes the coefficients obtained after inference. \emph{Operator inference}---the inference of coefficients over operator atoms---can be viewed as a gradient-based search over possible operator compositions. With the dictionary fixed, recurrent proximal-gradient steps minimize a layer-local energy that balances input reconstruction, sparsity, and agreement with a top-down signal. The support of $\vect{g}^{\star}$ identifies the atoms that \emph{participate}, and the coefficient values determine how strongly each contributes to the input-specific operator. During learning, the same coefficients gate \emph{plasticity}: only the parameter sets of the selected atoms are eligible for substantial updates. This selection identifies which operator-atom parameter sets participate in the current computation. It establishes functional parameter identifiability only if the same functional component remains linked to the same atom across contexts. Equation~\ref{eq:vn_effective_operator} describes the selected operator structure; the upward computation used by the model is defined below.

Within each layer, VNs compose selected operator atoms into an input-specific weight operator while allowing the upward propagated information state to remain dense. Across layers, these selected operators compose into the function required by the sample. Transformer attention changes how much each token uses other token states, while its learned projection matrices are shared across tokens. Inference in a VN instead selects parameterized operator atoms for the current sample, and the same selection determines which atoms' parameter sets may change.

In the rank-1 VN considered here, each operator atom has the form
\begin{equation}
\mat{A}_k = \vect{u}_k\vect{s}_k^\top,
\end{equation}
which maps a sensing direction $\vect{s}_k$ to an output direction $\vect{u}_k$. Let $\mat{S}=[\vect{s}_1,\ldots,\vect{s}_K]$ and $\mat{U}=[\vect{u}_1,\ldots,\vect{u}_K]$ collect the sensing and output vectors. The model uses the inferred coefficients to reconstruct the layer input and generate an upward message that can remain dense even when $\vect{g}^{\star}$ is sparse,
\begin{equation}
\widehat{\vect{x}}=\mat{S}\vect{g}^{\star},
\qquad
\vect{h}=\phi(\mat{U}\vect{g}^{\star}).
\end{equation}
where $\widehat{\vect{x}}$ is the reconstructed layer input, $\vect{h}$ is the dense upward message, and $\phi$ is the layer activation function. The upward message comes directly from the inferred code, rather than from applying the descriptive matrix in Equation~\ref{eq:vn_effective_operator} to $\vect{x}$. Inference and learning use the same layer-local energy,
\begin{equation}
E(\vect{g},\mat{S},\mat{U})=
\tfrac12\|\vect{x}-\mat{S}\vect{g}\|_2^2
+\lambda\|\vect{g}\|_1
+\tfrac12\|\vect{h}_{\mathrm{target}}-\mat{U}\vect{g}\|_2^2,
\label{eq:vn_layer_energy}
\end{equation}
where $E$ is the layer-local energy, $\vect{g}$ is the current participation or recruitment-coefficient vector, $\lambda$ controls recruitment sparsity, and $\vect{h}_{\mathrm{target}}$ is the top-down target supplied by the next layer. The top-down term is omitted when no such target is available. Fast operator inference holds $\mat{S}$ and $\mat{U}$ fixed and applies recurrent proximal-gradient steps to $\vect{g}$: the gradient step reduces the smooth reconstruction and top-down consistency terms, while proximal soft-thresholding enforces the $\ell_1$ sparsity penalty \citep{parikh2014proximal,beck2009fast}. This gradient-based search continues until the coefficients settle at $\vect{g}^{\star}$. Learning then holds those coefficients fixed and reduces the same energy by updating the atom vectors through local residuals,
\begin{equation}
\Delta\vect{s}_k\propto g_k^{\star}\vect{r}^{x},
\qquad
\Delta\vect{u}_k\propto g_k^{\star}\vect{r}^{h},
\label{eq:vn_hebb_error}
\end{equation}
where $\Delta\vect{s}_k$ and $\Delta\vect{u}_k$ denote updates to the atom vectors, $\vect{r}^{x}=\vect{x}-\mat{S}\vect{g}^{\star}$ is the input-reconstruction residual, and $\vect{r}^{h}=\vect{h}_{\mathrm{target}}-\mat{U}\vect{g}^{\star}$ is the top-down residual when a target is available. These factor updates preserve the rank-1 parameterization and are followed by normalization. Atoms with $g_k^{\star}\approx 0$ receive little or no error-driven update. The residuals depend on the joint explanation, so the selected atoms receive coordinated error-driven updates \citep{pokel2026learning}.

Rank-1 atoms are one choice of weight-operator granularity within the broader VN framework, not a general requirement. The key VN mechanism is the coupling through the energy: the same inferred code that determines which operator atoms participate also gates plasticity of their parameter sets. Because the parameter update uses local residuals at the settled coefficients, it does not differentiate through inference or backpropagate output derivatives through the full hierarchy. At deployment, the operator coefficients can be inferred from available observations and internal prediction signals without a supplied correct output.

The strongest result for VNs is evidence that a hierarchical model can organize the parameter sets of its rank-1 weight operators into a reusable factorization without manually assigning functional components to operators. Across controlled compositional benchmarks, VNs often achieve substantially lower (up to one order of magnitude) OOD error than dense baselines, including Transformer- and Mamba-style models (Figure~\ref{fig:weight_entanglement}d). This suggests that latent structure can be stored in reusable weight operators and that these operators can be composed into new functions, but does not yet show that individual operators implement the generating components, retain stable causal roles across contexts, or isolate the contribution of local learning. Those stronger claims require selective interventions, continual adaptation, and tasks whose factorization is not supplied by benchmark design.

\section{En route towards operator inference and functional parameter identifiability}
\label{sec:research_agenda}
The VN architecture first identifies which functional components must be selected anew for different samples or tasks. It represents this choice in a latent recruitment code, learns parameterized operators that implement reusable components, and infers the operator composition required for a new problem. Using the same weight operator recruitment code to gate learning then links each selected component to the parameter set of its operator and makes that set eligible for change. These requirements define a four-step framework for operator-based architectures:
\begin{center}
\fbox{\begin{minipage}{0.88\linewidth}
\begin{enumerate}[leftmargin=1.7em,itemsep=0.15em,topsep=0.3em]
    \item Identify which functional components or sub-functions should remain selectable across samples or tasks.
    \item Represent this selection with a sparse latent recruitment code $\vect{g}$.
    \item Learn weight operators that implement these reusable functional components.
    \item Infer and compose the operators required to form the function for each new problem.
\end{enumerate}
\end{minipage}}
\end{center}
From these requirements we derive four research priorities: establish stable operator roles, make recruitment efficient, extend functional parameter identifiability across depth and time, and benchmark compositional OOD generalization and selective change.

\paragraph{First, discover, stabilize and extend the reusable functional components.}
VNs show that rank-1 operator atoms can self-organize into reusable functional structure and be recruited in new compositions. A shared energy function used for inference links this recruitment to updates of the selected atoms' parameter sets, providing a concrete mechanism for functional parameter identifiability. The next step is to characterize this link at the level of individual atoms or stable atom compositions and determine how well it persists during continual learning. Selective ablations and targeted updates can test whether an operator continues to implement the same functional component across contexts and whether changing it leaves unrelated functions intact. Research should then determine which objectives or constraints best preserve these roles as the learned repertoire grows while retaining useful sharing. Stability concerns the functional component implemented by an operator and the transformation it performs; equivalent internal reparameterizations can preserve that role.

\paragraph{Second, make operator inference fast and efficient.}
Once operators have stable roles, the next challenge is to find the right ones quickly for each input. For reference, the Lottery Ticket Hypothesis shows that trainable sparse subnetworks can exist, but does not show how to find one on demand for each sample or task \citep{frankle2019lottery}. The analogous goal here is to “win the lottery” for every sample: infer a small composition of reusable weight operators that forms the function required for the current input, then update only the parameter sets of the selected operators when learning is needed. Such iterative inference can be slower than a feedforward pass, so a fast learned proposal could initialize recruitment before refinement against reconstruction, prediction, or consistency objectives. A router may speed the search, but its assignments need not make functional roles identifiable; both router-based and optimized selection must therefore be tested for stable operator roles. Comparisons should ask when refinement changes the selected operator composition, improves generalization, or enables more selective correction. They should also test whether using recent context or coarse-to-fine inference reduces the inference cost.

\paragraph{Third, extend functional parameter identifiability across depth and time.}
Efficient selection within one layer still leaves open whether functional components remain linked to identifiable parameter sets across layers and time. Across layers, a sample-specific function can depend on a chain of weight operators, each implementing a functional component that produces, transmits, or uses a representation. Identifying a feature at one layer does not identify this entire chain. Non-identifiability and basis changes at learned interfaces further complicate comparisons of internal structure \citep{dinh2017sharp,kornblith2019similarity,yosinski2014how,raghu2017svcca}. Entanglement arises when the relevant chain cannot be addressed selectively, rather than from depth itself. A hierarchy might first identify a relevant pathway or operator group, then identify the constituent operators that implement its functional components.

Functional parameter identifiability must also extend through time in recurrent systems, and the same operator may be used repeatedly before an outcome is known. A delayed outcome then raises the question of which use of which operator contributed to it and how local targets should guide correction. Selective correction may require addressing particular operator uses or compositions across a trajectory. Identifying the relevant operator uses within a trajectory differs from tracking how successive training phases modify the parameter set of the same operator; delayed credit assignment can remain difficult even when operator roles are well structured. Multimodal and embodied systems offer further tests across perception, world modeling, planning, and action.

Biological intelligence provides further inspiration for maintaining such identifiability across depth and time. Animals may rapidly recruit learned perceptual, motor, and cognitive structure in new combinations, while slower plasticity consolidates useful changes \citep{mathis2026leveraging}. This suggests testing architectures that compose operators rapidly across pathways and use local target-like and Hebbian-style signals to update the recruited operators during learning \citep{aceituno2025challenging}. The analogy motivates this design without assuming that the brain implements weight operators.

\paragraph{Fourth, benchmark compositional OOD generalization and selective change.}
Finally, diagnostic benchmarks should target compositional OOD cases with novel held-out combinations of factors that each appear during training, so the shift tests whether weight operators implementing familiar functional components can be recombined into a new function. Matched in-distribution cases should establish baseline performance. Evaluation should then ask whether held-out combinations reuse the operators associated with their familiar constituent components, and whether targeted edits or successive learning phases change the intended operators' parameter sets with limited collateral effects. The intervention criteria in Section~\ref{sec:functional_parameter_identifiability} provide measures of recruitment stability, selective deficits, collateral effects, and inference cost. Sparse routing may improve efficiency without functional parameter identifiability, so comparisons must separate gains from identifiable functional structure from gains due to extra capacity or inference.

Controlled experiments should establish whether a learned router or per-sample optimization selects the operators, and whether updates are limited to recruited operators or can also change unrecruited ones. The comparisons should use comparable architectures, data, and compute budgets. Systems should also be tested with the information available at deployment, including incomplete observations and uncertain targets. The decisive question is whether previously learned operators remain reusable, selectively accessible, and linked to the same functional components as situations change.

\section{Conclusion}
\label{sec:conclusion}
Robots, autonomous vehicles, and other systems operating in novel environments must reuse learned functional components in new combinations and adapt specific functions as conditions change. Their internal models should therefore make functional components identifiable in parameter space and selectively addressable. Weight entanglement describes the loss of a stable link between these functional components and identifiable parameter sets, with interference, difficult editing, and unreliable reuse as possible consequences. Existing forms of conditional computation provide a starting point, but functional parameter identifiability requires recruited operators to remain linked to the components they implement as those components change.

Research in real-world intelligent systems should thus prioritize architectures that infer compositions of reusable weight operators to form sample-specific functions and couple that inference to operator-gated learning. Comparisons with BP-trained baselines under matched size and update budgets, together with targeted ablations, should test what operator recruitment and local updates contribute. Evaluation should track operator recruitment across layers and time, measure whether targeted updates to the parameter sets of selected operators correct the intended component without disrupting unrelated functions, and report the cost of inference. The goal is an agent that can adapt rapidly in an open-ended world, reusing and composing familiar operators to project its knowledge onto new tasks, and updating only the parameter sets of the operators implementing the components that must change.

\section*{Acknowledgements}
We thank Sander de Haan and Yassine Taoudi-Benchekroun for comments on the manuscript. This work was supported by the Swiss National Science Foundation (189251, 10003772, B.F.G.) and ETH Zurich project funding (ETH-20 19-01, B.F.G.). The authors used OpenAI Codex and Anthropic Claude to assist with language editing and manuscript organization. The authors reviewed and approved all scientific arguments, interpretations, and final text.

\section*{Author contributions}
B.F.G. conceptualized the weight entanglement problem and wrote the original draft. G.C. and B.F.G. jointly revised and finalized the manuscript. Both authors approved the final version.

\section*{Competing interests}
The authors declare no competing interests.

\bibliographystyle{plainnat}
\bibliography{references}

\end{document}